\documentclass[twoside]{article}

\pdfoutput=1

\usepackage[accepted]{aistats2023}
\usepackage[round]{natbib}

\usepackage{float}
\usepackage{subcaption}
\usepackage{footmisc}
\usepackage[export]{adjustbox}
\usepackage{appendix}
\usepackage{amsmath}
\usepackage{array}
\usepackage[hyphens]{url}
\usepackage{enumitem}
\usepackage{xcolor,colortbl}

\begin{document}

\twocolumn[

\aistatstitle{Improving Expert Specialization in Mixture of Experts}

\aistatsauthor{ Yamuna Krishnamurthy \And Chris Watkins \And  Thomas G{\"a}rtner }

\aistatsaddress{ yamuna.krishnamurthy@rhul.ac.uk \\ Royal Holloway \\ Univerisity of London \And c.j.watkins@rhul.ac.uk \\ Royal Holloway \\ Univerisity of London \And  thomas.gaertner@tuwien.ac.at \\ TU Wien }

]

\begin{abstract}
 Mixture of experts (MoE), introduced over 20 years ago, is the simplest gated modular neural network architecture. There is renewed interest in MoE because the conditional computation allows only parts of the network to be used during each inference, as was recently demonstrated in large scale natural language processing models. MoE is also of potential interest for continual learning, as experts may be reused for new tasks, and new experts introduced. The gate in the MoE architecture learns task decompositions and individual experts learn simpler functions appropriate to the gate's decomposition. In this paper: (1) we show that the original MoE architecture and its training method do not guarantee intuitive task decompositions and good expert utilization, indeed they can fail spectacularly even for simple data such as MNIST and FashionMNIST; (2) we introduce a novel gating architecture, similar to attention, that improves performance and results in a lower entropy task decomposition; and (3) we introduce a novel data-driven regularization that improves expert specialization. We empirically validate our methods on MNIST, FashionMNIST and CIFAR-100 datasets.

\end{abstract}

\section{Introduction}
\label{sec:introduction}

Mixture of Experts (MoE) architecture was introduced by \cite{{jacobs1991a}} over $20$ years ago. It has since been successfully applied to learning problems such as reinforcement learning \citep{gimelfarb2018}, transfer learning \citep{mihai2017}, building large computationally efficient neural networks for language models and machine translation \citep{shazeer2017,rajbhandari2022, yazdaniaminabadi2022}, continual learning \citep{veniat2021,hihn2022} and learning multiple domains, such as image classification, machine translation, and image captioning, concurrently \citep{kaiser2017}.

MoE is a modular neural network architecture. They are the simplest and most successful modular neural network architectures. MoE consists of modules, called \textit{experts}, and a \textit{gate}. The experts and the gate are simple neural networks. The experts compute functions that are useful in different regions of the input space. The output of an expert, for each sample, is either the learnt class distribution for a classification problem or the learnt regression function output for a regression problem. For simplicity we will use classification problems in this paper.

\begin{figure}[H]
\centering
\includegraphics[width=1\linewidth]{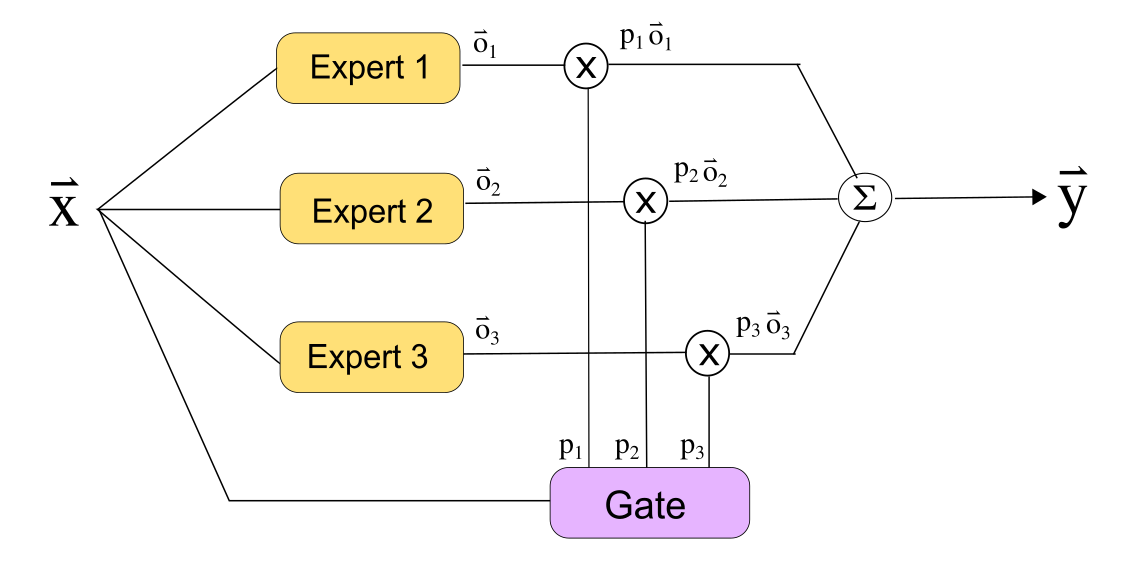}
\caption{Original Mixture of Experts (MoE) architecture with 3 experts and 1 gate. The output of the model is $\vec{\hat{y}} = p_1\cdot \vec{o}_1 + p_2\cdot \vec{o}_2 + p_3\cdot \vec{o}_3$, where $p_1$, $p_2$, $p_3$ are the gate outputs and $\vec{o}_1$, $\vec{o}_2$, $\vec{o}_3$ are the outputs of experts 1, 2 and 3 respectively.}
\label{fig:moe_expectation_arch} 
\end{figure}

The output of the gate is a vector of weights, one for each expert. The weights determine how much an expert contributes towards an MoE's prediction for a sample. This is called \textit{conditional computation} as only some experts are computed conditioned on the gate probabilities. Conditional computation is an important feature of an MoE as it makes training and inference faster. Ideally we want the gating network to learn a meaningful decomposition of the state space and the experts to learn simpler functions in different parts of the state space that results in better performance of the MoE model.

Figure \ref{fig:moe_expectation_arch} shows the \textit{output mixture model}, which is the original MoE architecture, introduced by \cite{jacobs1991a}. In this model the MoE prediction, $\vec{\hat{y}}$, is a weighted sum of the outputs of the experts, $\vec{\hat{y}} = \sum_{i=1}^{M}p_i\cdot \vec{o}_i$, where $\vec{o_i}$ is the output of expert $i$, $p_i$ is the gating weight for expert $i$ and $M$ is the number of experts. Since there are $M$ expert networks, the gating network has $M$ output units. The loss $L$ of the MoE is then $L = l(d,\vec{\hat{y}})$, where $d$ is the desired output and $l$ is a loss function. Since the output is a sum of proportions of the outputs of the experts, the experts are tightly coupled. The \textit{output mixture model} could seem to be not truly realizing conditional computation. In practice, however, the probabilities for some experts are small enough to be neglected. Those expert outputs need not be computed and so indeed does enable conditional computation.





MoE models are of particular interest because of their:

\begin{enumerate}[noitemsep]
\item{faster training due to conditional weight updates and faster inference due to conditional computation during feed forward \citep{shazeer2017},}
\item{transferability of sub-tasks learnt by experts to other tasks \citep{mihai2017}. This makes them especially attractive to continual learning \citep{veniat2021,hihn2022},}
\item{parallelizable expert training \citep{rajbhandari2022},}
\item{ability to solve  multi-modal problems with a combination of heterogeneous experts \citep{kaiser2017}.}
\item{ability to solve multi-task problems with multi-gate MoE architectures \citep{ma2018}.}
\end{enumerate}

The current literature on MoE, however, has concentrated on the performance of the overall model and not on what each expert learns. Our first contribution is our finding and clear presentation of two crucial problems in training MoE models: (1) that original MoE training methods lead to inequitable and unintuitive task decompositions that have both poor error and loss; and (2) how the tasks are distributed among the experts is relevant to both their performance and scalability. 

Our second contribution is a novel MoE gating architecture, we call \textit{attentive gating architecture}. In current MoE, the expert distribution by the gate for a sample does not depend on the computations of the experts on that sample. This is to allow conditional computation, however, it seems unreasonable for the gate to learn the task decomposition by itself. Both the expert and gate learn sample classification and expert distribution, respectivley, based on the same input distribution. It then seems intuitively reasonable to not duplicate this learning. The attentive gating architecture computes the gate's expert distribution for a given sample, as the attention score, computed with the gate and expert computations for the given sample. The proposed method is analogous to computing the self-attention score, proposed by \cite{bahdanau2015}, of the gate and expert outputs. This is effectively asking the question, \textbf{Which experts should the gate attend to for a given sample?}


Our experiments show that the attentive gating approach results in lower entropy of the task decomposition without compromising performance. However, since the task decomposition depends on expert computations there is no conditional computation during feed forward when training. We show that we can still provide conditional computation during inference by distilling the model, trained with attentive gating, to the original MoE architecture with no loss in performance.
    
MoE trains both experts and the gate `end-to-end` on overall loss of the model. The training does not provide any incentive for equitable use of experts, that is, a more balanced and intuitive distribution of samples across experts. We observed that this results in some experts being starved of samples during training. The starved experts, that are not allocated any or very few samples during training, are effectively not used for inference. An extreme version of this is when the gate selects the same expert for all the samples. \cite{kirsch2018} refer to this as \textit{module collapse}. When module collapse occurs, the MoE output does not depend on the gate. This is equivalent to using a single model.

Our third and last contribution addresses this problem with a data-driven constraint, $L_s$, added as a regularization term to the loss. $L_s$ routes the samples that are similar, determined by a similarity measure, to the same expert and those that are not to different experts. In our experiments we have used the Euclidean distance as the dissimilarity measure (it is a dissimilarity measure because a larger distance indicates dissimilarity). The method could use other (dis)similarity measures. We have not tested any other measures.


Our paper is organised as follows: Section \ref{sec:related_work} discusses the related work; Section \ref{sec:metrics} defines the information theoretic performance metrics we use to analyse the performance of the different MoE models and their training methods that we use in this paper; Section \ref{sec:equitable_decomposition} presents the results of our preliminary experiments to analyse how a task is distributed by the gate among the experts. The findings in this section are our first contribution; Section \ref{sec:attentive_gating} introduces our second contribution, a novel \textit{attentive gating MoE architecture}; Section \ref{sec:sample_similarity} introduces our third contribution, a novel data-driven soft constraint regularizatoin, $L_s$; Section \ref{sec:experiments} details our experiments with the novel attentive gating architecture and $L_s$ regularization and presents their results; we finally conclude with Section \ref{sec:conclusion}. Our repository\footnote{\url{https://github.com/aistats2023-1554/moe}} has the code to reproduce all the experiments and results reported in this paper.


\section{Related Work}
\label{sec:related_work}

\textbf{Expert specialization in MoE:} Much of the MoE research so far has concentrated on the performance of the MoE model and not on how the task is decomposed between the experts. Recently there has been interest in improving the expert specialization through improved task decomposition as it improves performance and conditional computation \citep{shazeer2017}. In \cite{mittal2022} the authors have performed similar experiments as us to compare the specialization of experts trained `end-to-end' with those trained with a good task decomposition. They arrived at the same conclusion that the original MoE training methods indeed lead to poor expert specialization and that a good task decomposition results in better expert specialization and better performance. Our work, presented as our first contribution, however pre-dates theirs as it was presented in our earlier work at a NeuRIPS 2021 workshop [citation hidden for anonymity]. \cite{mittal2022} evaluated using synthetic data while we have used real data to arrive at the same results.

\textbf{Task specific expert specialization:} There have been quite a few approaches to task specific expert specializations, especially for language and vision tasks, recently \citep{kudugunta2021, riquelme2021, lewis2021, lepikhin2021, fedus2022, zhou2022}. In all these approaches routing decisions to experts are based on image and text tokens. Hence, they are task aware approaches where the experts have to be a specific architecture and can only work with one type of dataset. So they are not well suited for multi-modal learning. For example, \cite{riquelme2021, lepikhin2021, fedus2022} have added sparsity to transformer architectures by using MoE in the dense network layer of transformers for vision and language. Our approach is task agnostic. Each of our experts could have a different architecture.

\textbf{Expert specialization with regularization:} Since the `end-to-end' MoE training provides no incentive for an equitable sample distribution to the experts, auxiliary losses were added as regularizations by \cite{shazeer2017, lewis2021}. The regularization added by \cite{lewis2021} is specific to their method of routing text tokens to the experts. \cite{shazeer2017} proposed a more generic $L_{importance}$ regularization for equitable task distribution. However, as discussed in Section \ref{sec:sample_similarity}, their method simply uses all available experts even when it is not required for the task. Our regularization, $L_s$ discussed in Section \ref{sec:sample_similarity}, is a data-driven approach to equitable task distribution that is a more scalable solution. Their work is the most relevant to ours.

\textbf{Attentive gating:} To the best of our knowledge our attentive gate architecture is novel. The only other related work we found was by \cite{liu2020}, who have used the attention mechanism in the gate to focus the gate on different aspects of the input and target images. The gate then learns good segmentation of the input images and assigns the different segments to different experts. Their approach is similar to the original MoE where the gate independently decides the tasks to be assigned to the experts by attending to the data. Our approach learns the gate's expert distribution by attending to the experts.



\section{Information Theoretic Performance Metrics}
\label{sec:metrics}

Accuracy or error is not sufficient to measure the performance of an MoE as we are also interested in measuring gating sparsity and expert usage. We will here define the information theoretic performance metrics we use to analyse the training of the MoE. These metrics measure how well the gate distributes the samples to the experts and how well it utilizes the experts. 

\subsection{Measuring Gating Sparsity}

Conditional computation is an important feature of the MoE. Sparser gating probabilities are desirable because they result in better conditional computation. The sparsity per sample can be measured by the average per sample expert selection entropy, $H_s$, in Equation \ref{eq:h_s}, over a batch. $N$ is the number of samples in a batch and $\vec{p}=\left(p_1,p_2\ldots p_M\right)$ are the gate probabilities for $M$ experts, for each sample. A low value of $H_s$ indicates sparse gating probabilites and hence better conditional computation.
\begin{align}
    H_s & = \frac{1}{N} \sum_{i=1}^{N}{H(\vec{p}_i)} 
\label{eq:h_s}
\end{align}

\begin{figure*}[h]
\begin{subfigure}[t]{0.5\textwidth}
\centering 
\includegraphics[width=1\linewidth]{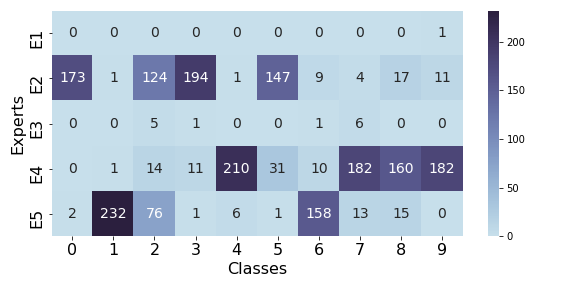}
\subcaption{MNIST}
\label{fig:mnist_exp_tab_scratch}
\end{subfigure}%
\begin{subfigure}[t]{0.5\textwidth}
\centering 
\includegraphics[width=1\linewidth]{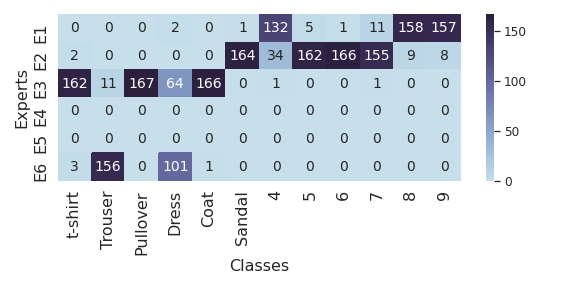}
\subcaption{FMNIST and MNIST}
\label{fig:fmnist_mnist_exp_tab_scratch}
\end{subfigure}
\caption{Expert selection table of the original MoE model for: (a) MNIST and (b) combined FMNIST and MNIST, datasets. We can see that not all experts are used. The task decomposition is not intuitive as in the case of combined FMNIST and MNIST expert 2 is used for both FMNIST and MNIST classes. }
\end{figure*}

\subsection{Measuring Expert Utilization}

Ideally we want the sub-tasks of the task to be distributed equitably between the experts to avoid module collapse. This will require the average gate probabilities for each of the experts, over all the samples, to be roughly equal. The distribution of the experts over the samples can be measured by the entropy of the average gate probabilities over all samples in a batch, $H_u$, as in Equation \ref{eq:h_u}. A high $H_u$ indicates a more equitable gate probability distribution and hence better utilization of experts. A low $H_u$ indicates unequal utilization of experts. For example, in the case of module collapse, when all samples get sent to the same expert, that expert's average gate probability is $1.0$. The probabilities of all the other experts will be zero. This will result in $H_u=0$. 

\begin{equation}
    H_u = H\:\Biggl(\frac{1}{N} \sum_{i=1}^{N}{\vec{p}_i}\:\Biggr)   
\label{eq:h_u}
\end{equation}


\subsection{Measuring model output dependency on expert selection}

We introduce a new metric to measure the dependency of the class distribution $Y$ on the gate's expert selection distribution $E$. An equitable gate task decomposition among experts results in a high mutual dependence between $Y$ and $E$.

In the case of module collapse, one expert does all the work and the gate does not contribute to solving the task. There is then no dependency between $E$ and $Y$. In the case where each expert is assigned just one sub-task, the gate does all the work. There is then a higher dependency between $E$ and $Y$. Hence, the more equitable the task distribution between the experts the higher the dependence between $E$ and $Y$.

The mutual dependency between $E$ and $Y$ can be measured by computing their mutual information, I(E;Y), as shown in Equation \ref{eq:mutual}, where $H(E)$ is the marginal entropy of $E$, $H(Y)$ is the marginal entropy of $Y$ and $H(E,Y)$ is the joint entropy of $E,Y$. Higher $I(E;Y)$ values indicate better dependence between $E$ and $Y$ and subsequently more equitable task decomposition.

\begin{equation}
I(E;Y) \equiv H(E)+H(Y)-H(E,Y)
\label{eq:mutual}
\end{equation}

Since we do not have the true marginal and joint probabilities of $E$ and $Y$, we compute them empirically as detailed in Appendix \ref{app:mutual_info}. The sample sizes are large enough that we do not introduce significant estimation bias. 


\begin{figure*}[h]
\begin{subfigure}[t]{0.5\textwidth}
  \includegraphics[width=1\linewidth]{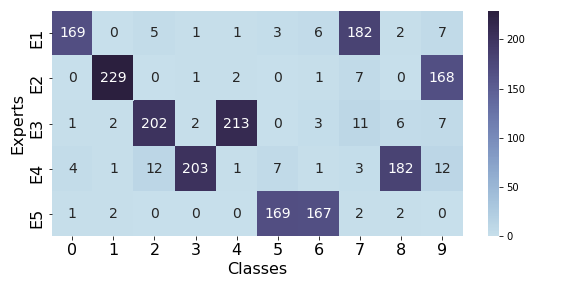}
  \subcaption{MNIST}
  \label{fig:mnist_exp_tab_prechosen}
 \end{subfigure}%
 \begin{subfigure}[t]{0.5\textwidth}
  \includegraphics[width=1\linewidth]{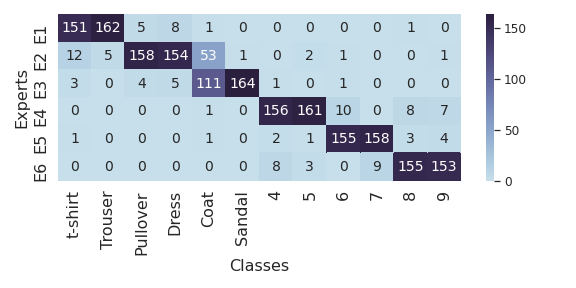}
  \subcaption{FMNIST and MNIST}
  \label{fig:fmnist_mnist_exp_tab_prechosen}
 \end{subfigure}
\caption{Expert selection table of models trained with experts pre-trained on custom splits of the classes: (a) MNIST: \{[0,7], [1,9], [2,4], [3,8], [5,6]\} and (b) combined FMNIST and MNIST: \{[t-shirt,Trouser], [Pullover,Dress],[Coat,Sandal],[4,5],[6,7],[8,9]\}}
\label{fig:fmnist_mnist_prechosen}
\end{figure*}

\section{Better Performance with Better Expert Specialization}
\label{sec:equitable_decomposition}

We will now look more closely at what the experts in an MoE model learn and show that the original MoE training approaches cannot find intuitive task decompositions. This results in poor expert specialization with a few experts learning most of the task. We show that intuitive and balanced task decompositions are important because they lead to better expert specialization which in turn improves the performance of MoE models.

\subsection{Does original MoE training find intuitive task decompositions?}
\label{sec:moe_inequitable}

We ran preliminary experiments to analyse how the gate, in the original MoE, distributes a classification task among the experts. Our experiments showed that the original MoE model does not find a balanced and intuitive task decomposition and hence expert usage, even for the simple MNIST \citep{lecun2010} learning problem. We trained an MoE model, that has $5$ experts and $1$ gate, on $10,000$ training samples of the MNIST data containing all the $10$ digits. We chose $5$ experts as the MNIST dataset has $10$ classes (sub-tasks). This allows for an intuitive distribution of $2$ classes per expert.


Each expert and the gate is a simple convolutional network with a single convolutional layer and 2 hidden layers with ReLU activation. For details of the parameters of the model please refer to Appendix \ref{app:mnist}. We trained with Adam optimizer.

The trained model was used to classify $2,000$ samples of the MNIST test data. Figure \ref{fig:mnist_exp_tab_scratch} is an expert selection table. Each cell of the table is the count of samples of the digit that were routed to the expert corresponding to the cell. Figure \ref{fig:mnist_exp_tab_scratch} shows that only $3$ of the $5$ experts are used.  

Since the MNIST dataset contains only digits, we thought we should try with a dataset that contains clearly very different sets of images with the intuition that samples from different datasets would be routed to different experts. We created such a dataset by combining the FashionMNIST (FMNIST) \citep{xiao2017} and MNIST datasets. We chose the first $6$ classes, $[t-shirt,\ trouser,\ pullover,\ dress,\ coat,\ sandal]$, from FMNIST and last $6$ classes,$[4,\ 5,\ 6,\ 7,\ 8,\ 9]$, from MNIST and combined the data to create one dataset of $12$ classes. The model for combined FMNIST and MNIST dataset has $6$ experts as there are $12$ classes, with the expert and gate architectures same as the MNIST model. For details of the parameters of the model refer to Appendix \ref{app:mnist+fmnist}.

Figure \ref{fig:fmnist_mnist_exp_tab_scratch} shows that the gate surprisingly uses expert $2$ to learn a mix of classes from FMNIST and MNIST. Hence, we see that an intuitive task decomposition in MoE is not guaranteed even in a seemingly trivial case where the images of FMNIST and MNIST are clearly quite different from each other. In Section \ref{sec:equitable_better} we also see that such decompositions not only use experts inequitably but also result in poor performance. Let us now analyse the possible reasons for such unintuitive task decompositions. 


\subsection{Do intuitive task decompositions have better performance?}
\label{sec:equitable_better}



The simplest method to train an MoE is to train the gate and experts at the same time, `end-to-end', by gradient descent. During training the gating probabilities, for each sample, determine which experts get trained on that sample. That is, gating interacts with training and in effect experts are trained only when they are chosen by the gating network. Existing MoE architectures trained `end-to-end' do not decompose the task intuitively among the experts as we saw in Section \ref{sec:moe_inequitable}. The question we are trying to answer is: does the `end-to-end' MoE training find a gating decomposition that performs well for the task, even though it seems surprisingly counter-intuitive? Or, is the search for gating decomposition simply bad?

\begin{figure}[H]
\centering
\includegraphics[width=0.9\linewidth]{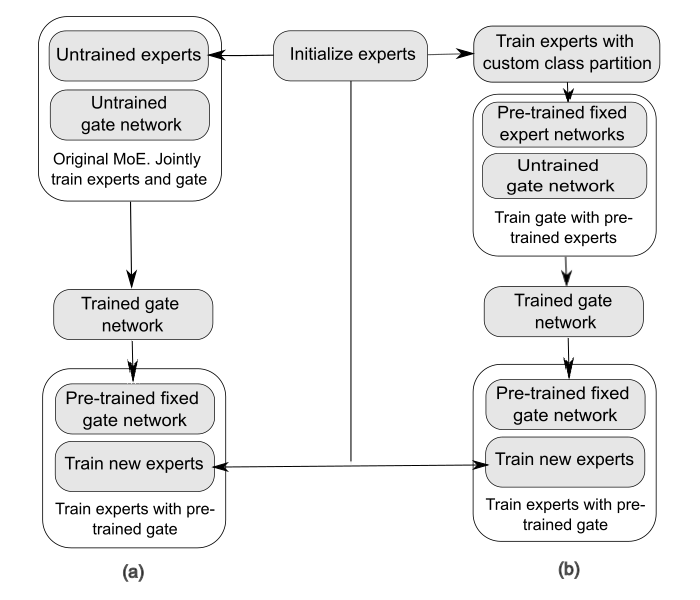}
\caption{Experiment designed to analyse if intuitive task decompositions have better performance. Refer to the Table \ref{tab:fmnist_pretrain_results} for results of the experiment.}
\label{fig:exp_design} 
\end{figure}

We designed an experiment, summarized in Figure \ref{fig:exp_design}, to answer these questions. What we need for this is: (1) a gate trained with un-trained experts, using the original MoE model, resulting in unintuitive task decomposiiton as in Section \ref{sec:moe_inequitable}; and (2) a gate trained with experts pre-trained with custom intuitively plausible partitions of the dataset. We then use each of these two pre-trained gates to train a new set of experts with the same decomposition of the task as the experts the gates were trained with. This enables us to check the performance of the gate task decompositions for an unintuitive partition vs an intuitive partition. 

Firstly, let us define a more intuitive task decomposition for the MNIST dataset and determine if the gate can learn this decomposition. We split the $10$ digits into $5$ sets of $5$ pairs of digits, such as $\{[0,7], [1,9], [2,4], [3,8], [5,6]\}$. We used $5$ experts, each of which was trained with only data samples of one of the $5$ pairs of digits. So the pairs of digits are distributed equally among the experts.

We then fixed the parameters of these pre-trained experts and trained the gate with them. From the gate expert selection table in Figure \ref{fig:mnist_exp_tab_prechosen}, we see that the gate can indeed learn to select the correct expert for each digit and hence learn an intuitive task decomposition. Figure \ref{fig:fmnist_mnist_exp_tab_prechosen} shows the gate expert selection table for one split of the combined FMNIST and MNIST dataset, trained in the same way as with the MNIST dataset. We again see that the gate can learn to select the correct expert for each class in the combined dataset.

We then fixed the parameters of the pre-trained gate and trained the MoE model with the pre-trained gate and new experts. Both the pre-trained gates decomposed the tasks exactly as in Figures \ref{fig:mnist_exp_tab_scratch} and \ref{fig:mnist_exp_tab_prechosen} respectively for the MNIST dataset and similarly as Figures \ref{fig:fmnist_mnist_exp_tab_scratch} and \ref{fig:fmnist_mnist_exp_tab_prechosen} for the combined FMNIST and MNIST dataset. Hence we see that a gate can learn an intuitive task decomposition.

Let us now check the training loss and test error of the models with intuitive and unintuitive task decompoitions. Tables \ref{tab:mnist_pretrain_results} and \ref{tab:fmnist_pretrain_results} show the average training loss and average test error, both averaged over $5$ runs of the experiment for MNIST and combined FMNIST and MNIST datasets. We see that the model trained with pre-trained experts has a lower training loss than the model trained with un-trained experts and has a lower error rate for both datasets.

\begin{table}[H]
\centering
\caption{Comparison of average training loss and test error for MoE models: (a) with inequitable task decompositions; and (b) with equitable task decompositions, from the experiment detailed in Figure \ref{fig:exp_design}, for MNIST and combined MNIST and FMNIST datasets.}
\small
\begin{minipage}{.5\linewidth}
\begin{tabular}{|m{1cm}| m{1cm} | m{0.7cm} |}
\hline
\textbf{Models}&\textbf{Test Error} & \textbf{Train Loss} \\ \hline
(a) & 0.12 & 0.19\\ \hline
(b) & 0.08 & 0.05 \\ \hline
\end{tabular}
\subcaption{MNIST}
\label{tab:mnist_pretrain_results}
\end{minipage}%
\begin{minipage}{.5\linewidth}
\begin{tabular}{|m{1cm}| m{1cm} | m{0.7cm}|}
\hline
\textbf{Models}&\textbf{Test Error} & \textbf{Train Loss} \\ \hline
(a) & 0.15 & 0.28 \\ \hline
(b) & 0.10 & 0.13 \\ \hline
\end{tabular}
\subcaption{MNIST and FMNIST}
\label{tab:fmnist_pretrain_results}
\end{minipage}
\end{table}

\vspace{-0.5cm}

The experiment shows that intuitive task decompositions do exist with much better performance. The gate, however, does not learn them when both experts and the gate are jointly trained `end-to-end'. The gate initially finds a poorly performing and unintuitve task decomposition and reinforces that throughout the training. If we have prior knowledge of a good task decomposition then it would be best to pre-train the experts on these sub-tasks and then train the gate. Typically we do not know a plausible task decomposition and it is what we wish to find, but `end-to-end' MoE training fails to do so, even in this simple case.




\section{Attentive Gating MoE Architecture}
\label{sec:attentive_gating}

\begin{figure}[H]
\centering
\includegraphics[width=1.0\linewidth]{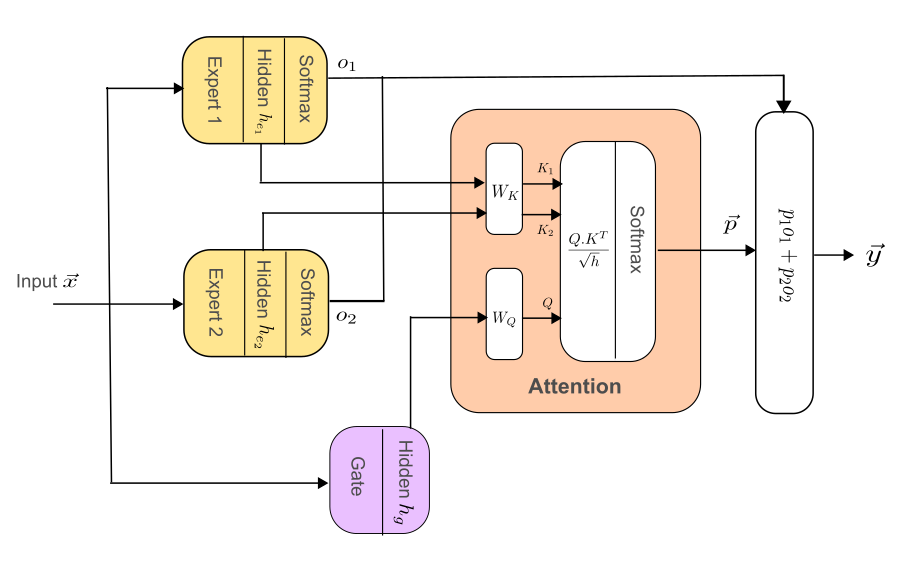}
\caption{Attentive gating MoE architecture.}
\label{fig:attentive_gate_arch}
\end{figure}

In current MoE the gate learns the expert distribution from the input distribution and the expert learns the classification of the samples based on the input and expert distribution by `end-to-end` training.  We suggest a more intuitively plausible design, shown in Figure \ref{fig:attentive_gate_arch}, that uses the expert's computations in computing the gating distribution.


During MoE training, the gate output is the current query or token of interest and the expert outputs are the sequence of tokens that are attended to. The gate's hidden output, $G_{1\times h}$ (subscripts are the size of the matrix), is used to compute the \textit{Query}, $Q_{1\times h}$, as in Equation \ref{eq:query} and the expert hidden outputs, $E_{i_{1\times h}}$, are used to compute the \textit{Keys}, $K_{i_{1\times h}}$, as in Equation \ref{eq:keys}, where $E_i$ is the $i^{th}$ expert of $M$ experts in the model. $h$ is the size of the hidden layers of the experts and the gate. $W_{q_{h\times h}}$ and $W_{k_{h\times h}}$ are the query and key weight matrices.

The attention score $A(Q, K)$ (we have dropped the subscripts of $Q$ and $K$ here for better readability) is then computed as in Equation \ref{eq:attention_score}:

\vspace{-0.7cm}
\begin{align}
    Q_{1\times h} &= G_{1\times h}\cdot W_{q_{h\times h}}
    \label{eq:query}\\
    K_{i_{1\times h}} &= E_{i_{1\times h}}\cdot W_{k_{h\times h}}
    \label{eq:keys}\\
    A(Q_{1\times h}, K_{M\times h}) &= softmax\Biggl(\frac{Q_{1\times h}\cdot K_{M\times h}^T}{\sqrt{h}}\Biggr)
    \label{eq:attention_score}
\end{align}

The computed attention $A(Q,K)$ can then be used to weight the outputs of the experts. Hence, $A(Q,K)$ are the gate probabilities of selecting the corresponding expert, to compute the MoE output and loss.

Our experiments, in Section \ref{sec:experiments}, show that with the attentive gate the MoE model performs better than the original MoE method but does have a similar problem of inequitable expert utilization. Hence, there is a need for a soft constraint that will ensure equitable expert utilization. We discuss the regularization we used to tackle this problem in Section \ref{sec:sample_similarity}.

\subsection{Distilling attentive gating MoE model for conditional computation}
\label{sec:attn_inference}

In the attentive gate architecture gating is dependent on the expert computations during feed forward. This does not allow for conditional computation during inference. To address this we distill the MoE model, trained in Section \ref{sec:attentive_gating}, into a regular MoE \textit{output mixture model}. We fix the parameters of the experts learnt using the attentive gate, initialise the gate of the new \textit{output mixture model} to the trained gate parameters and proceed to train the new MoE model and gate.


\section{Gating with Sample Similarity Regularization}
\label{sec:sample_similarity}

We need a soft constraint to ensure an equitable sample distribution to the experts. \cite{shazeer2017} proposed the $L_{importance}$ loss regularization as a soft constraint to assign equal importance to all experts for a batch. $L_{importance}$ measures the batch-wise coefficient of variation (CV) of the gate output probabilities to avoid module collapse as in Equation \ref{eq:loss_importance}. $\vec{I}{=}\sum_{x\in X}{\vec{p}_x}$ is an importance factor that measures the relative importance of the expert to the batch with $X$ samples. $\vec{p}_x$ is the gate's expert distribution for sample $x \in X$. $w_{importance}$ is a tunable hyperparameter. $CV(\vec{I}){=}\sigma(\vec{I})/\mu(\vec{I})$, where $\sigma$ is the standard deviation and $\mu$ is the mean. 

\begin{equation}
    L_{importance}\left(X\right) = w_{importance} \cdot CV(\vec{I})
\label{eq:loss_importance}
\end{equation}

The  $L_{importance}$ regularization, however, just aims at using all the experts available equally and not in a suitable way for the task. This results in poor scalability as we show in Section \ref{sec:experiments}. It seems intuitive and natural to add a data-driven soft constraint based on the properties of the samples in the dataset. This would allow incorporating domain knowledge into the training. Samples belonging to the same task tend to be similar. The hypothesis here is that, routing similar samples to the same expert and dissimilar samples to different experts will ensure cleaner and more equitable task decomposition.


With this in mind we propose a data-driven soft constraint by adding a regularization factor, $L_s$, based on some similarity measure of the samples.

\begin{align}
\large
   L_{s}(X) &= \frac{1}{(N^2-N)}\Bigl[\sum_{x,x'} S(x,x') - D(x,x')\Bigr] \\
   S(x,x') &= \frac{1}{M}\sum_{e}{\beta_{s}\cdot p(e|x)\cdot p(e|x')\cdot \Vert x-x'\Vert^2} \\
   D(x,x') &= \frac{1}{(M^2-M)}\sum_{e\neq e'}{\beta_{d}\cdot p(e|x)\cdot p(e'|x')\cdot\Vert x-x'\Vert^2}
\normalsize                     
\label{eq:sample_similarity}
\end{align}

We have used the squared Euclidean distance measure $\Vert x-x'\Vert^2$, for pairs of samples $x,x' \in X$, where $X$ is a batch of size $N$. The purpose of the regularization is to allow the gate to learn expert selection probabilities, $p(e|x)$, for each sample such that it minimizes the term, $S(x,x')$, with similar samples routed to the same expert and maximises the term, $D(x,x')$, with dissimiar samples sent to different experts as in Equation \ref{eq:sample_similarity}, where $M$ is the number of experts in the model, $e,e' \in E_M$ are the experts assigned to samples $x,x'$ respectively and $\beta_s$, $\beta_d$ are tunable hyperparameters.

Our experitments detailed in Section \ref{sec:experiments} show that $L_s$ regularization performs as well as or better than $L_{importance}$ regularization, while using less experts.


\section{Experiments}
\label{sec:experiments}

We evaluate our methods on the small MNIST and the much larger CIFAR-100 \citep{krizhevsky09} datasets. For the MNIST dataset we used an MoE model with $5$ experts and $1$ gate. For the the CIFAR-100 dataset we used $20$ experts and $1$ gate.

Each expert for the MNIST dataset has: $1$ convolutional layer; $2$ hidden layers with $ReLU$ activation; and one output layer. The gate has the same architecture as the expert but different parameters. For details of the parameters of the model refer to Appendix \ref{app:mnist}. 


Each expert for the CIFAR-100 dataset has: $4$ convolutional layers; We used batch normalization and max pooling layers; $2$ hidden layers with $ReLU$ activation; and one output layer. For details of the parameters of the model refer to Appendix \ref{app:cifar100}.

All models were trained with Adam optimizer with $0.001$ learning rate. We used $20$ epochs for MNIST dataset and $40$ epochs for CIFAR-100 dataset. Each experiment was run $10$ times for MNIST dataset and $5$ times for CIFAR-100 dataset. 

Our baseline for the MoE architecture is the original MoE architecture and training method, the output mixture model. Our baseline for MoE regularization is the $L_{importance}$ \citep{shazeer2017} regularization which is a generic regularization. Other MoE regularizations in the literature are specific to certain architectures and training methods.


We trained the models for both datasets as follows: (1) single model which has the same architecture as one expert; (2) vanilla or original MoE \textit{output mixture model} with no regularizations; (3) vanilla MoE with $L_{importance}$ regularization with different values of  $w_{importance}$; (4) vanilla MoE with $L_s$ regularization with different combinations of values of $\beta_s$ and $\beta_d$; (5) with attentive gating MoE architecture; (6) with attentive gating MoE and $L_{importance}$ regularization for different values of $w_{importance}$; (7) with attentive gating and $L_s$ regularization for different combinations of values of $\beta_s$ and $\beta_d$; (8) model distilled from attentive gating MoE with $L_{importance}$; and (9) model distilled from attentive gating MoE with $L_s$. The values of all the hyperparameters used in the experiments are listed in Appendix \ref{app:hyper}.

The experiment results for MNIST dataset are in Table \ref{tab:mnist_results}. The experiment results for CIFAR-100 dataset are in Table \ref{tab:cifar100_results}. The results for each method of training, in the tables, are the performance metrics computed on the test set, with the the model that has the minimum training error among the multiple runs for each method. The standard deviation of the test error over the runs is also reported.

\begin{table}[h!]
\centering
\caption{Performance on the test set of the model with the minimum training error for MNIST dataset. Best results in each category of MoE training approaches is highlighted.}
\small
\begin{tabular}{|m{2.5cm}|m{1.6cm}|m{0.9cm}|m{0.6cm}|m{0.6cm}|}
\hline
\textbf{Experiment}&\textbf{Error} & $\mathbf{I(E;Y)}$ & $\mathbf{H_s}$ & $\mathbf{H_u}$ \\ \hline
single model & 0.096$\pm$0.071 & NA & NA &  NA\\ \hline \hline
vanilla MoE & 0.038$\pm$0.009 & 2.022 & 0.092 & 2.172 \\ \hline 
vanilla MoE with $L_{importance}$ & 0.032$\pm$0.008 & 2.262 & 0.061 & 2.32\\ \hline
\textbf{vanilla MoE with} $\mathbf{L_s}$ & \textbf{0.029}$\mathbf{\pm}$\textbf{0.009} & 2.244 & 0.051 & 2.246 \\ \hline \hline
attentive gate MoE & 0.033$\pm$0.006 & 1.797 & 0.071 & 2.055 \\ \hline
attentive gate MoE with $L_{importance}$ & 0.035$\pm$0.005 & 2.26 & 0.055 & 2.266 \\ \hline
\textbf{attentive gate MoE with} $\mathbf{L_s}$ & \textbf{0.032}$\mathbf{\pm}$\textbf{0.006} & 2.275 & 0.039 & 2.321 \\ \hline \hline
distilled from attentive gate MoE with $L_{importance}$ & 0.030$\pm$0.007 & 2.301 & 0.036 & 2.32 \\ \hline
\textbf{distilled from attentive gate MoE with} $\mathbf{L_s}$ & \textbf{0.028}$\mathbf{\pm}$\textbf{0.007} & 2.191 & 0.056 & 2.319 \\
\hline
\end{tabular}
\label{tab:mnist_results}
\end{table}

\begin{table}[h!]
\centering
\caption{Performance on the test set of the model with the minimum training error for CIFAR-100 dataset. Best results in each category of MoE training approaches is highlighted.}
\small
\begin{tabular}{|m{2.5cm}|m{1.6cm}|m{0.9cm}|m{0.6cm}|m{0.6cm}|}
\hline
\textbf{Experiment}&\textbf{Error} & $\mathbf{I(E;Y)}$ & $\mathbf{H_s}$ & $\mathbf{H_u}$ \\ \hline
single model & 0.575$\pm$0.006 & NA & NA &  NA\\ \hline \hline
vanilla MoE & 0.460$\pm$0.010& 0.967 & 0.306 & 1.023 \\ \hline
vanilla MoE with $L_{importance}$ & 0.483$\pm$0.007& 4.177 & 1.135 & 3.981 \\ \hline
\textbf{vanilla MoE with} $\mathbf{L_s}$ & \textbf{0.457}$\mathbf{\pm}$\textbf{0.012}& 1.424 & 0.381 & 1.279 \\ \hline \hline
attentive gate MoE & 0.450$\pm$0.006& 1.792 & 0.463 & 2.178 \\ \hline
\textbf{attentive gate MoE with $L_{importance}$} & \textbf{0.447}$\pm$\textbf{0.005}& 3.684 & 1.036 & 4.141 \\ \hline
attentive gate MoE with $\mathbf{L_s}$ & 0.451$\mathbf{\pm}$0.016& 3.117 & 0.770 & 3.357 \\ \hline \hline
distilled from attentive gate MoE with $L_{importance}$ & 0.531$\pm$0.131 & 3.179 & 1.75 & 3.843 \\ \hline
\textbf{distilled from attentive gate MoE with} $\mathbf{L_s}$ & \textbf{0.482}$\mathbf{\pm}$\textbf{0.065} & 1.605 & 0.718 & 2.541 \\
\hline
\end{tabular}
\label{tab:cifar100_results}
\end{table}

Tables \ref{tab:mnist_results} and \ref{tab:cifar100_results} show that the attentive gating model performs better than original MoE. Combined training with attentive gate and $L_{importance}$ or $L_s$ regularizations improves expert usage as indicated by higher $H_u$ values and improves gate sparsity as indicated by lower $H_s$ values.


We also see that the $L_s$ regularization has lower error rate than $L_{importance}$. $L_s$ does as well as or better than $L_{importancee}$ in terms of expert usage with higher values of $H_u$. $L_s$ regularization also has better conditional inference due to lower $H_s$.

We also evaluated with the FMNIST dataset. The details and results for the FMNIST dataset are in Appendix \ref{app:fmnist} and \ref{app:fmnist_results}.

\begin{figure}[H]
\centering
\begin{subfigure}[b]{0.5\linewidth}
\includegraphics[width=1\linewidth]{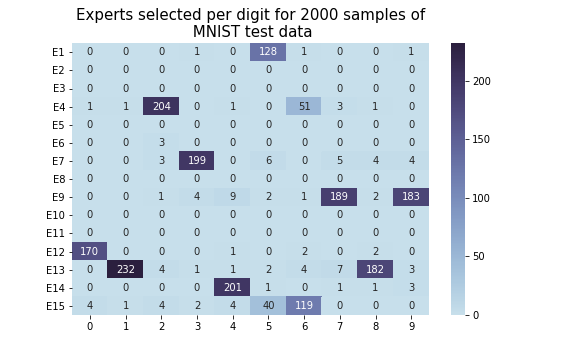}
\subcaption{$L_s$}
\end{subfigure}%
\begin{subfigure}[b]{0.5\linewidth}
\includegraphics[width=1\linewidth]{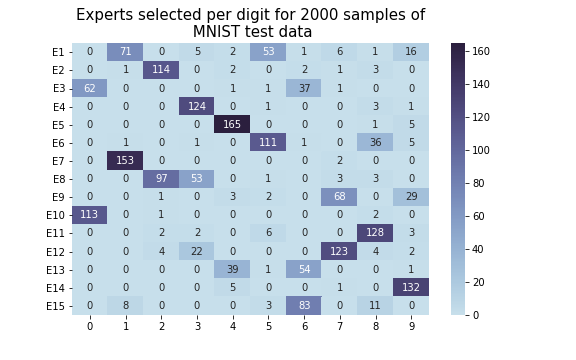}
\subcaption{$L_{importance}$}
\end{subfigure}
\caption{Expert selection table of MoE model trained with $L_s$ and $L_{importance}$ regularizations with $15$ experts.}
\label{fig:10_15_experts}
\end{figure}

Another discernible improvement of $L_s$ over $L_{importance}$ is in the number of experts required for the task. $L_{importance}$ is designed to use all the available experts equitably whether this is required for the task or not. We ran experiments by increasing the number of experts from $5$ to $15$ for the MNIST dataset, which is more than the  number of classes for the MNIST dataset. Figure \ref{fig:10_15_experts} shows that $L_s$ regularization results in more optimal use of experts while $L_{importance}$ uses all the experts. This implies models with $L_s$ could have less parameters than with $L_{importance}$. Results with $10$ experts are in Appendix \ref{app:10_10_experts}.

\section{Conclusion}
\label{sec:conclusion}

In this paper we have clearly shown that intuitive task decompositions by the gate perform better. We introduced a novel MoE model architecture and training method using attentive gating. This method of training computes the gate's expert distribution on a sample from the computations of the experts for that sample. Finally, we introduce a novel data-driven sample similarity regularization, $L_s$, that distributes the samples between the experts based on sample similarity. Our experiments show that training with attentive gating and $L_s$ regularization improves performance, expert specialization and gate sparsity.

\bibliography{moe_bibliography}

\section*{Appendix}

\appendix

\section{Computing Mutual Information Between Class Distribution, $Y$, and the Gate's Expert Distribution, $E$}
\label{app:mutual_info}

Mutual information, $I(E;Y)$, between the class distribution $Y$ and the gate expert distribution $E$, can be computed by first computing the count $c_{ij}$, where $i\in\{1\dots, M\}$ and $j\in\{1,\dots,K\}$, $M$ is the number of experts and $K$ is the number of classes in the task. $c_{ij}$ is the number of times expert $E_i$ is selected for samples of class $Y_j$. $c_{ij}$ is computed for each expert, for each class and hence we have an $M\times K$ count matrix $C$. 

\begin{table}[h!]
\centering
\caption{Matrix $C$ of count of number of times $E_i$ is selected for class $Y_j$}
\begin{tabular}{|c|c|c|c|}
\hline
Count(E,Y) & $Y_1$ & \dots & $Y_K$ \\
\hline
$E_1$ &$c_{11}$ &\dots & $c_{1K}$\\
\hline
$\vdots$ & $\vdots$ & $\ddots$ & $\vdots$ \\
\hline
$E_M$ &$c_{M1}$ & \dots & $c_{MK}$ \\
\hline
\end{tabular}
\label{tab:count_matrix}
\end{table}

 From $C$, we can compute the batchwise joint and marginal probabilities of $E,Y$ in Table \ref{tab:e_y_prob} using Equations \ref{eq:joint_ey}, \ref{eq:marginal_e} and \ref{eq:marginal_y}, where $N$ is the total number of samples in a batch:
 
Joint Probabilty $P(E,Y)$:
\begin{equation}
P(E{=}E_i,Y{=}Y_j) =p(E_i,Y_j) =c_{ij}/N  
\label{eq:joint_ey} 
\end{equation}

Marginal Probability $P(E)$:
\begin{equation}
P(E{=}E_i) = p(E_i) = \sum_{j=1}^K{p(E_i,Y_j)} 
\label{eq:marginal_e} 
\end{equation}        

Marginal Probability $P(Y)$:
\begin{equation}
P(Y{=}Y_j) = p(Y_j) = \sum_{i=1}^M{p(E_i,Y_j)} 
\label{eq:marginal_y} 
\end{equation}                 
 
\begin{table}[h!]
\centering
\caption{Joint and marginal probabilities of $E$ and $Y$}
\small
\begin{tabular}{|c|c|c|c|c|}
\hline
$P(E,Y)$& $Y_1$ & \dots & $Y_K$ & $P(E_i)$\\
\hline
$E_1$ &$p(E_1,Y_1)$ &\dots & $p(E_1,Y_K))$ & $p(E_1)$ \\
\hline
$\vdots$ & $\vdots$ & $\ddots$ & $\vdots$ & $\vdots$ \\
\hline
$E_M$ & $p(E_M,Y_1)$ & \dots & $p(E_M,Y_K)$ & $p(E_M)$ \\
\hline
$P(Y_j)$ & $p(Y_1)$  & $\dots$ & $p(Y_K)$ & 1 \\
\hline
\end{tabular}
\label{tab:e_y_prob}
\end{table}

We can now compute the required entropies in Equation \ref{eq:mutual} from quantities computed in Table \ref{tab:e_y_prob} using Equations \ref{eq:entropy_e}, \ref{eq:entropy_y}, \ref{eq:entropy_ey}. Subsequently we can compute the mutual information $I(E;Y)$ as in Equation \ref{eq:mutual}.


\begin{align}
    H(E) &= \sum_{i=1}^M{-p(E_i)\log_2\, p(E_i)}
    \label{eq:entropy_e}\\
    H(Y) &= \sum_{j=1}^K{-p(Y_j)\log_2\, p(Y_j)}
    \label{eq:entropy_y} \\
    H(E,Y) &= \sum_{i=1}^M\sum_{j=1}^K{-p(E_i,Y_j)\log_2\, p(E_i,Y_j)}
\label{eq:entropy_ey}
\end{align}

\section{Neural Network Architecture and Parameter Details for MoE Models in Experiments}
\label{app:model_details}

The number of experts in the MoE model, for the MNIST and FMNINST datasets, is half the number of classes in the dataset. For example, MoE model for MNIST has $5$ experts as MNIST has $10$ classes. We chose experts to be half the number of classes for the dataset as this allows for an equitable distribution of $2$ classes per expert (all datasets used in the paper have even number of classes).

For CIFAR-100 dataset we chose $20$ experts as there are $20$ coarse labels. The total unique class labels are $100$, grouped into $20$ coarse labels each with $5$ classes. Hence $20$ experts allows for equitable class distribution among experts if required. In practice, however, one could experiment with different number of experts to determine the optimal model for a given task. All the MoE models used in the paper have one $1$ gate. 

For each dataset we tried different expert and gate architectures and parameters for the original MoE model. We then chose the MoE model with minimum train error, for the given dataset. We used the same expert and gate architectures and parameters of the selected MoE model for all the training methods on that dataset.

We used PyTorch for our implementation. All experiments were run on a single GPU.  

\subsection{MoE model for MNIST dataset}
\label{app:mnist}

MoE model for MNIST dataset has $5$ experts and $1$ gate. The training set has $60,000$ samples, the test set has $10,000$ samples. There are $10$ classes.

\begin{description}
\item[Expert:] Each expert of all the MoE models has $1$ convolutional layer, $2$ hidden layers and $1$ output layer. The details of the layers are as follows:
\begin{itemize}
\item $1$ convolutional layer with $1$ input channel, $1$ output channel and a kernel size of $3$, with \textit{ReLU} activation and max pooling with kernel size $2$ and stride $2$,
\item $2$ hidden layers with $ReLU$ activation. First  hidden layer has input of $1*13*13$ and output of $5$. Second hidden layer has input of $5$ and output of $32$,
\item $1$ output layer with input $32$ and output $10$, which is the number of classes, with \textit{ReLU}activation and 
\item softmax layer
\end{itemize}

\item[Original Gate:] The gate for the original MoE model has $1$ convolutional layer, $2$ hidden layers and $1$ output layer. The details of the layers are as follows:
\begin{itemize}
\item $1$ convolutional layer with $1$ input channel, $1$ output channel and a kernel size of $3$, with \textit{ReLU} activation and max pooling with kernel size $2$ and stride $2$,
\item $2$ hidden layers with $ReLU$ activation. First  hidden layer has input of $1*13*13$ and output of $128$. Second hidden layer has input of $128$ and output of $32$,
\item $1$ output layer with input $32$ and output $5$, which is the number of experts, with \textit{ReLU}activation
\item softmax layer
\end{itemize}

\item[Attentive Gate:] The attentive gate has $1$ convolutional layer and $2$ hidden layers. The details of the layers are as follows:
\begin{itemize}
\item $1$ convolutional layer with $1$ input channel, $1$ output channel and a kernel size of $3$, with \textit{ReLU} activation and max pooling with kernel size $2$ and stride $2$. This is the same as the original gate,
\item $2$ hidden layers. First  hidden layer has input of $1*13*13$ and output of $128$ with $ReLU$ activation, this is the same as the orginal gate for the first hidden layer. Second hidden layer has input of $128$ and output of $32$ and no activation. This is the output of the attentive gate used to compute the query and attention score.
\item There are no output and softmax layers.
\end{itemize}

\end{description}

\subsection{MoE model for combined FashionMNIST (FMNIST) and MNIST dataset}
\label{app:mnist+fmnist}

MoE model for combined FNIST and MNIST dataset has $6$ experts and $1$ gate. The training set has $10,000$ samples and the test set has $2,000$ samples. We chose the first $6$ classes, \textit{[t-shirt, trouser, pullover, dress, coat, sandal]}, from FMNIST and last $6$ classes,\textit{[4, 5, 6, 7, 8, 9]}, from MNIST and combined the data to create one dataset of $12$ classes.

\begin{description}
\item[Expert:] Each expert, of all the MoE models for the combined FMNIST and MNIST dataset, has the same architecture and parameters as that for the MNIST dataset in Appendix \ref{app:mnist}. Only the output layer output is $12$ as the combined FMNIST and MNIST dataset has $12$ classes.

\item[Original Gate:] The gate for the  MoE model has the same architecture as that for the MNIST dataset in Appendix \ref{app:mnist}, but with different parameters.
\begin{itemize}
\item $1$ convolutional layer with $1$ input channel, $1$ output channel and a kernel size of $5$, with \textit{ReLU} activation and max pooling with kernel size $2$ and stride $2$,
\item $2$ hidden layers with $ReLU$ activation. First  hidden layer has input of $1*12*12$ and output of $128$. Second hidden layer has input of $128$ and output of $32$,
\item $1$ output layer with input $32$ and output $6$, which is the number of experts, with \textit{ReLU}activation
\item softmax layer
\end{itemize}

\end{description}

\subsection{MoE model for CIFAR-100  dataset}
\label{app:cifar100}

MoE model for CIFAR-100 dataset has $20$ experts and $1$ gate. The training set has $50,000$ samples and the test set has $10,000$ samples. There are $100$ unique classes.

\begin{description}
\item[Expert:] Each expert of all the MoE models for the CIFAR-100 dataset has $4$ convolutional layers, $2$ hidden layers and $1$ output layer. The details of the layers are as follows:
\begin{itemize}
\item $1$ convolutional layer with $3$ input channels, $16$ output channels and a kernel size of $3$, with \textit{ReLU} activation and max pooling with kernel size $2$ and stride $2$,
\item $1$ convolutional layer with $16$ input channels, $32$ output channels and a kernel size of $3$, with batch normalization with $32$ features, with \textit{ReLU} activation and max pooling with kernel size $2$ and stride $2$,
\item $1$ convolutional layer with $32$ input channels, $64$ output channels and a kernel size of $3$, with \textit{ReLU} activation and max pooling with kernel size $2$ and stride $2$,
\item $1$ convolutional layer with $64$ input channels, $128$ output channels and a kernel size of $3$, with batch normalization with $128$ features, with \textit{ReLU} activation and max pooling with kernel size $2$ and stride $2$,  
\item $2$ hidden layers with $ReLU$ activation. First  hidden layer has input of $128*2*2$ and output of $1024$. Second hidden layer has input of $1024$ and output of $256$,
\item $1$ output layer with input $256$ and output $100$, which is the number of classes, with \textit{ReLU}activation and 
\item softmax layer
\end{itemize}

\item[Original Gate:] The gate for the original MoE model has $4$ convolutional layers, $2$ hidden layers and $1$ output layer. The details of the layers are as follows:
\begin{itemize}
\item $1$ convolutional layer with $3$ input channel, $64$ output channel and a kernel size of $3$, with \textit{ReLU} activation and max pooling with kernel size $2$ and stride $2$,
\item $1$ convolutional layer with $64$ input channel, $128$ output channel and a kernel size of $3$, with batch normalization with $128$ features, with \textit{ReLU} activation and max pooling with kernel size $2$ and stride $2$,
\item $1$ convolutional layer with $128$ input channel, $256$ output channel and a kernel size of $3$, with \textit{ReLU} activation and max pooling with kernel size $2$ and stride $2$,
\item $1$ convolutional layer with $256$ input channel, $512$ output channel and a kernel size of $3$, with batch normalization with $512$ features, with \textit{ReLU} activation and max pooling with kernel size $2$ and stride $2$,  
\item $2$ hidden layers with $ReLU$ activation. First  hidden layer has input of $512*2*2$ and output of $1024$. Second hidden layer has input of $1024$ and output of $256$,
\item $1$ output layer with input $256$ and output $20$, which is the number of experts, with \textit{ReLU}activation and 
\item softmax layer
\end{itemize}

\item[Attentive Gate:] The attentive gate has $4$ convolutional layers and $2$ hidden layers. The details of the layers are as follows:
  \begin{itemize}
\item $1$ convolutional layer with $3$ input channel, $64$ output channel and a kernel size of $3$, with \textit{ReLU} activation and max pooling with kernel size $2$ and stride $2$,
\item $1$ convolutional layer with $64$ input channel, $128$ output channel and a kernel size of $3$, with batch normalization with $128$ features, with \textit{ReLU} activation and max pooling with kernel size $2$ and stride $2$,
\item $1$ convolutional layer with $128$ input channel, $256$ output channel and a kernel size of $3$, with \textit{ReLU} activation and max pooling with kernel size $2$ and stride $2$,
\item $1$ convolutional layer with $256$ input channel, $512$ output channel and a kernel size of $3$, with batch normalization with $512$ features, with \textit{ReLU} activation and max pooling with kernel size $2$ and stride $2$,  
\item $2$ hidden layers with $ReLU$ activation. First  hidden layer has input of $512*2*2$ and output of $1024$. Second hidden layer has input of $1024$ and output of $256$,
\item There are no output and softmax layers.
\end{itemize}

\end{description}

\subsection{MoE model for FMNIST  dataset}
\label{app:fmnist}

MoE model for FMNIST dataset has $5$ experts and $1$ gate. The training set has $60,000$ samples and the test set has $10,000$ samples. There are $10$ classes.

\begin{description}
\item[Expert:] Each expert of all the MoE models for the FMNIST dataset has the same architecture as that of the MNIST dataset in Appendix \ref{app:mnist} but more parameters in the hidden layers. The FashionMNIST is more complex than the MNIST digits data and hence needs more parameters:
\begin{itemize}
\item $1$ convolutional layer with $1$ input channel, $1$ output channel and a kernel size of $3$, with \textit{ReLU} activation and max pooling with kernel size $2$ and stride $2$,
\item $2$ hidden layers with $ReLU$ activation. First  hidden layer has input of $1*13*13$ and output of $64$. Second hidden layer has input of $64$ and output of $32$,
\item $1$ output layer with input $32$ and output $10$, which is the number of classes, with \textit{ReLU}activation and 
\item softmax layer
\end{itemize}

\item[Original Gate:] The gate for the original MoE model has the same architecture as that of the MNIST dataset in Appendix \ref{app:mnist} but more filters in the output channels and more parameters in the hidden layers.
\begin{itemize}
\item $1$ convolutional layer with $1$ input channel, $8$ output channels and a kernel size of $3$, with \textit{ReLU} activation and max pooling with kernel size $2$ and stride $2$,
\item $2$ hidden layers with $ReLU$ activation. First  hidden layer has input of $8*13*13$ and output of $512$. Second hidden layer has input of $512$ and output of $32$,
\item $1$ output layer with input $32$ and output $5$, which is the number of experts, with \textit{ReLU}activation
\item softmax layer
\end{itemize}

\item[Attentive Gate:] The attentive gate has $1$ convolutional layer and $2$ hidden layers. The details of the layers are as follows:
\begin{itemize}
\item $1$ convolutional layer with $1$ input channel, $8$ output channel and a kernel size of $3$, with \textit{ReLU} activation and max pooling with kernel size $2$ and stride $2$. This is the same as the original gate,
\item $2$ hidden layers. First  hidden layer has input of $8*13*13$ and output of $512$ with $ReLU$ activation. Second hidden layer has input of $512$ and output of $32$ and no activation. This is the output of the attentive gate used to compute the query and attention score.
\item There are no output and softmax layers.
\end{itemize}

\end{description}

\section{Hyperparameter Values Used for Experiments}
\label{app:hyper}

In our experiments we trained each model with different values of the corresponding hyperparameters. We then chose the model with the lowest training error for each category of the model and training methods. The hyperparameters we tuned are $w_{importance}$ for $L_{importance}$ regularization and $\beta_s$ and $\beta_d$ for $L_s$ regularization.

The values used for the $w_{importance}$ hyperparameter of the $L_{importance}$ regularization, for all datasets, are $w_{importance}{=}\{0.2, 0.4, 0.6, 0.8, 1.0\}$.

The values used for $\beta_s$ and $\beta_d$ hyperparameters of the $L_s$ regularization, for different datasets are summarized in Table \ref{tab:betas}

\begin{table}[h!]
\centering
\caption{Values of hyperparameters (H) $\beta_s$ and $\beta_d$ for datasets (D).}
\small
\begin{tabular}{|m{1.5cm}| m{2.2cm} | m{3.2cm} |}
\hline
\textbf{D/H}& $\mathbf{\beta_s}$ & $\mathbf{\beta_d}$ \\ \hline
MNIST & $\{1\text{e-}6,1\text{e-}5\}$ & $\{10^{-i}\mid i\in\{1,\ldots,6\}\}$\\ \hline
FMNIST & $\{1\text{e-}7,1\text{e-}6\}$ & $\{10^{-i}\mid i\in\{1,\ldots,7\}\}$\\ \hline
CIFAR-100 & $\{1\text{e-}5,1\text{e-}4,1\text{e-}3\}$ & $\{10^{-i}\mid i\in\{1,\ldots,7\}\}$\\ \hline
\end{tabular}
\label{tab:betas}
\end{table}

\begin{figure*}[h]
\centering
\begin{subfigure}[b]{0.5\textwidth}
\includegraphics[width=1\linewidth]{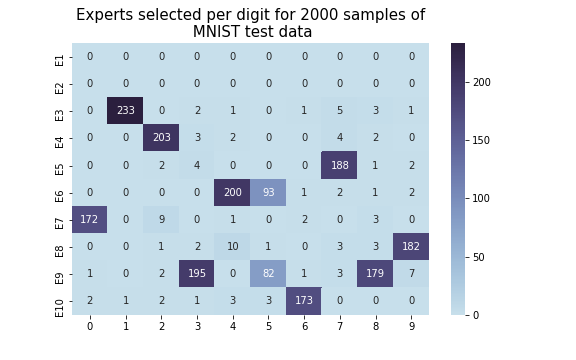}
\subcaption{$L_s$}
\label{fig:importance_reg_10_experts}
\end{subfigure}%
\begin{subfigure}[b]{0.5\textwidth}
\includegraphics[width=1\linewidth]{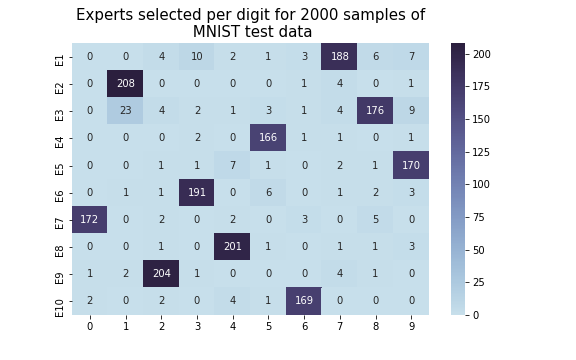}
\subcaption{$L_{importance}$}
\label{fig:sample_sim_reg_10_experts}
\end{subfigure}
\caption{Expert selection table of MoE model trained with $L_s$ and $L_{importance}$ regularizations with $10$ experts.}
\label{fig:10_10_experts}
\end{figure*}

\section{Results for FMNIST dataset}
\label{app:fmnist_results}

\begin{table}[h!]
\centering
\caption{Performance on the test set of the model with the minimum training error for FMNIST dataset. Best results in each category of MoE training approaches is highlighted.}
\small
\begin{tabular}{|m{2.8cm}|m{1.6cm}|m{0.9cm}|m{0.6cm}|m{0.6cm}|}
\hline
\textbf{Experiment}&\textbf{Error} & $\mathbf{I(E;Y)}$ & $\mathbf{H_s}$ & $\mathbf{H_u}$ \\ \hline
single model & 0.132$\pm$0.011 & NA & NA &  NA\\ \hline \hline
vanilla MoE & 0.104$\pm$0.006& 2.034 & 0.081 & 2.048 \\ \hline
vanilla MoE with $L_{importance}$ & 0.103$\pm$0.006& 2.301 & 0.172 & 2.321 \\ \hline
\textbf{vanilla MoE with} $\mathbf{L_s}$ & \textbf{0.095}$\mathbf{\pm}$\textbf{0.007}& 2.198 & 0.11 & 2.225 \\ \hline \hline
attentive gate MoE & 0.098$\pm$0.013& 2.071 & 0.114 & 2.249 \\ \hline
attentive gate MoE with $L_{importance}$ & 0.098$\pm$0.006& 2.233 & 0.101 & 2.319 \\ \hline
\textbf{attentive gate MoE with} $\mathbf{L_s}$ & \textbf{0.096}$\mathbf{\pm}$\textbf{0.008}& 2.296 & 0.109 & 2.321 \\ \hline \hline
\textbf{distilled from attentive gate MoE with} $\mathbf{L_{importance}}$ &\textbf{0.087}$\mathbf{\pm}$\textbf{0.007}& 2.227 & 0.101 & 2.318 \\ \hline
distilled from attentive gate MoE with $L_s$ & 0.089$\pm$0.008& 2.304 & 0.076 & 2.321 \\
\hline
\end{tabular}
\label{tab:fmnist_results}
\end{table}

Refer to Appendix \ref{app:fmnist} for details of the MoE architecture for the FMNIST dataset. Table \ref{tab:fmnist_results} shows the results for FMNIST dataset.

Table \ref{tab:fmnist_results} again shows that MoE model with attentive gating performs the best. Also $L_s$ performs better than $L_{importance}$ regularization. The distilled attentive MoE models perform the best.

\section{Expert Usage on As Needed Basis}
\label{app:10_10_experts}

In Section \ref{sec:experiments} we showed that when we use $15$ experts for $10$ classes in the MNIST dataset, the $L_{importance}$ regularization uses all the experts whereas the $L_s$ regularization uses as many experts as is required for the task. We also increased the number of experts from $5$ to $10$ which is the same number of experts as the number of classes in MNIST. Figure \ref{fig:10_10_experts} shows that $L_s$ regularization results in more optimal use of experts while $L_{importance}$ uses all the experts.

\end{document}